\documentclass[11pt]{article}

\usepackage[final]{acl}

 \usepackage{microtype}

\usepackage[english]{babel} 
\babelfont{rm}{TeXGyreTermesX} 
\babelprovide[import]{russian}
\babelfont[russian]{rm}[Extension=.otf, UprightFont=Tempora-Regular, BoldFont=Tempora-Bold, ItalicFont=Tempora-Italic, BoldItalicFont=Tempora-BoldItalic]{Tempora} 
\babelfont{tt}[Extension=.otf, UprightFont=FreeMono, BoldFont=FreeMonoBold, ItalicFont=FreeMonoOblique, BoldItalicFont=FreeMonoBoldOblique]{FreeMono} 

\usepackage{graphicx}
\usepackage{booktabs}
\usepackage{multirow}
\usepackage{amsmath}
\usepackage{listings}
\usepackage{enumitem}
\usepackage{fvextra}
\usepackage{tikz}
\usetikzlibrary{arrows.meta, positioning, fit, calc}

\title{Learning Nested Named Entity Recognition from Flat Annotations}

\author{Igor Rozhkov \\
  Lomonosov Moscow State University \\ Leninskie Gory, 1/4, Moscow, Russia \\
  \texttt{fulstocky@gmail.com} \\\And
  Natalia Loukachevitch \\
  Lomonosov Moscow State University \\ Leninskie Gory, 1/4, Moscow, Russia \\
  \texttt{louk\_nat@mail.ru} \\}

\begin{document}

\maketitle
\begin{abstract}
Nested named entity recognition identifies entities contained within other entities, but requires expensive multi-level annotation. While flat NER corpora exist abundantly, nested resources remain scarce. We investigate whether models can learn nested structure from flat annotations alone, evaluating four approaches: string inclusions (substring matching), entity corruption (pseudo-nested data), flat neutralization (reducing false negative signal), and a hybrid fine-tuned + LLM pipeline. On NEREL, a Russian benchmark with 29 entity types where 21\% of entities are nested, our best combined method achieves 26.37\% inner F1, closing 40\% of the gap to full nested supervision. Code is available at \url{https://github.com/fulstock/Learning-from-Flat-Annotations}.
\end{abstract}

\section{Introduction}

Named entity recognition (NER) is a fundamental task in information extraction \citep{tjong-kim-sang-de-meulder-2003-introduction,ratinov-roth-2009-design,lample-etal-2016-neural}. While conventional NER systems assume non-overlapping entity spans, real-world text frequently contains nested entities where one mention is contained within another. For example, in ``Ministry of Foreign Affairs of the United Kingdom,'' the entity ``United Kingdom'' (COUNTRY) is nested within the larger ORGANIZATION. Recognizing such nested structures can benefit downstream tasks \citep{finkel-manning-2009-nested,lu-roth-2015-joint}: relation extraction may leverage more complete entity information, entity linking---disambiguate entities at multiple nesting levels, and knowledge graph construction can capture more finer-grained entity relationships.

However, nested NER requires expensive annotation where annotators must identify entity mentions at every level of nesting. Early nested NER datasets include ACE 2004 \citep{doddington-etal-2004-automatic} and GENIA \citep{kim-etal-2003-genia}, with larger-scale resources appearing more recently \citep{ringland-etal-2019-nne,loukachevitch-etal-2021-nerel}. The annotation cost creates a practical barrier: creating nested annotations requires much more effort than flat annotation. Furthermore, recent attempts to use large language models for automatic nested annotation have shown limited success, with LLMs struggling to maintain consistency across nesting levels \citep{kim-etal-2024-exploring}.

In contrast, flat NER corpora---marking only non-overlapping entities---exist abundantly across languages and domains. Years of NER research have produced flat datasets for dozens of languages, various domains (news, biomedical, legal, social media), and different entity type systems. This abundance motivates our research questions:

\begin{itemize}[itemsep=2pt,topsep=2pt]
    \item \textbf{RQ1:} Can we train models to recognize nested entities using only flat annotations?
    \item \textbf{RQ2:} Can large language models help bridge the gap between flat and nested supervision?
    \item \textbf{RQ3:} Can large language models compensate for missing nested annotations?
\end{itemize}

\noindent If successful, these approaches would enable leveraging existing flat NER resources for nested recognition without additional annotation effort.

We investigate these questions using NEREL \citep{loukachevitch-etal-2021-nerel}, a Russian nested NER benchmark with 29 entity types---significantly more complex than typical nested NER datasets with 4--8 types. In NEREL, 21\% of all entity mentions are nested within other entities. We systematically compare methods for recovering nested structure from flat annotations:

\begin{enumerate}[itemsep=2pt,topsep=2pt]
    \item \textit{Inclusions}: identifying nested entities through substring matching across the corpus.
    \item \textit{Entity corruption}: creating pseudo-nested training data by corrupting tokens within entities.
    \item \textit{Flat neutralization}: reducing false negative signal from unlabeled nested positions.
    \item \textit{Hybrid fine-tuned + LLM}: using fine-tuned models for outer entities and LLMs for nested detection.
\end{enumerate}

\begin{figure*}[t]
  \centering
\resizebox{\textwidth}{!}{%
\begin{tikzpicture}[
    node distance=0.3cm and 0.4cm,
    box/.style={draw, rounded corners=2pt, minimum height=0.65cm, align=center, font=\footnotesize},
    databox/.style={box, fill=blue!8, minimum width=1.4cm},
    methodbox/.style={box, fill=orange!12, minimum width=1.5cm, minimum height=0.75cm},
    modelbox/.style={box, fill=green!10, minimum width=1.4cm},
    outputbox/.style={box, fill=gray!12, minimum width=1.3cm},
    exbox/.style={draw, rounded corners=2pt, fill=yellow!8, font=\scriptsize,
                  align=left, inner sep=2.5pt},
    groupbox/.style={draw, dashed, rounded corners=4pt, inner sep=5pt, gray!70},
    arr/.style={-{Stealth[length=4pt]}, semithick},
    panellabel/.style={font=\footnotesize\bfseries},
    glabel/.style={font=\scriptsize\itshape, text=gray!70!black}
]


\node[panellabel] (title-a) at (0, 0) {(a) Fine-tuned Approaches};

\node[databox, below=1.1cm of title-a.south west, anchor=north west] (nerel)
  {NEREL\\[-2pt]{\scriptsize(flat)}};

\node[methodbox, right=1.05cm of nerel, yshift=0.55cm] (incl) {Inclusions};
\node[methodbox, right=1.0cm of nerel, yshift=-0.55cm] (corr) {Entity\\[-2pt]Corruption\\[-2pt]{\scriptsize(5-fold retrain)}};

\node[methodbox, right=1.0cm of nerel, xshift=2.4cm] (neut)
  {Flat\\[-2pt]Neutral.};

\node[modelbox, right=0.6cm of neut] (binder) {Binder\\[-2pt]{\scriptsize(fine-tuned)}};

\node[outputbox, right=0.6cm of binder] (pred-a) {Nested\\[-2pt]predictions};

\coordinate (fork) at ([xshift=0.35cm]nerel.east);
\draw[semithick] (nerel.east) -- (fork);
\draw[arr] (fork) |- (incl.west);
\draw[arr] (fork) |- (corr.west);

\coordinate (join) at ([xshift=-0.35cm]neut.west);
\draw[semithick] (incl.east) -| (join);
\draw[semithick] (corr.east) -| (join);
\draw[arr] (join) -- (neut.west);

\draw[arr] (neut) -- (binder);
\draw[arr] (binder) -- (pred-a);

\node[groupbox, fit=(incl)(corr),
      label={[glabel, anchor=south]above:data augmentation}] {};
\node[groupbox, fit=(neut),
      label={[glabel, anchor=south]above:training modification}] {};

\node[exbox, text width=5.0cm,
      below=1.0cm of nerel.south west, anchor=north west,
      label={[font=\scriptsize\bfseries, inner sep=1pt, xshift=-1.0cm]above left:Inclusions}
      ] (ex-incl) {%
  ``Min.\ of Foreign Affairs of \underline{Russia}'' {\tiny(ORG.)}%
  \\$\rightarrow$ ``Russia'' {\tiny(COUNTRY)} added as inner%
};

\node[exbox, text width=5.0cm,
      below=0.45cm of ex-incl.south west, anchor=north west,
      label={[font=\scriptsize\bfseries, inner sep=1pt, xshift=-0.2cm, yshift=-0.05cm]above left:Entity Corruption}
      ] (ex-corr) {%
  ``Min.\ of Foreign Affairs of \textit{klr}''%
  \\$\rightarrow$ predicts ``Min.\ of Foreign Affairs'' {\tiny(ORG.)}%
};

\node[exbox, text width=5.2cm, anchor=west,
      label={[font=\scriptsize\bfseries, inner sep=1pt, xshift=0.25cm]above left:Flat Neutralization}
      ] (ex-neut)
  at ([xshift=0.25cm]$(ex-incl.east)!0.5!(ex-corr.east)$) {%
  ``\underline{Min.\ of Foreign Affairs} of \underline{Russia}'' {\tiny(ORG)}%
  \\[1pt]{\tiny\textbf{+}} ``Min.\ of Foreign Affairs of Russia'' \hfill {\tiny positive}%
  \\{\tiny\textbf{--}} ``of Foreign'', ``Affairs of'', \ldots \hfill {\tiny negative}%
  \\{\tiny$\circ$} ``Russia'' {\tiny(matches COUNTRY)} \hfill {\tiny neutral}%
  \\[1pt]$\rightarrow$ neutral spans excluded from loss%
};


\node[panellabel, below=0.5cm of ex-corr.south west, anchor=north west] (title-b)
  {(b) LLM-based Approaches};

\node[font=\scriptsize\itshape, below=0.3cm of title-b.south west, anchor=north west] (pure-label) {Pure LLM:};

\node[databox, right=0.15cm of pure-label.east, anchor=west, minimum width=1.1cm] (input1)
  {Input text};
\node[modelbox, right=2.3cm of input1, fill=purple!10] (llm1)
  {LLM\\[-2pt]{\scriptsize(few-shot)}};
\node[outputbox, right=2.35cm of llm1] (out1)
  {All entities\\[-2pt]{\scriptsize(outer+inner)}};

\draw[arr] (input1) -- (llm1);
\draw[arr] (llm1) -- (out1);

\node[font=\scriptsize\itshape, below=0.8cm of pure-label.south west, anchor=north west] (hybrid-label) {Hybrid:};

\node[databox, right=0.45cm of hybrid-label.east, anchor=west, minimum width=1.1cm] (input2)
  {Input text};
\node[modelbox, right=0.2cm of input2, fill=green!10] (binder2)
  {Binder};
\node[outputbox, right=0.2cm of binder2, minimum width=1.2cm] (outer2)
  {Outer\\[-2pt]entities};
\node[modelbox, right=0.2cm of outer2, fill=purple!10] (llm2)
  {LLM\\[-2pt]{\scriptsize(per span)}};
\node[outputbox, right=0.2cm of llm2, minimum width=1.0cm] (inner2)
  {Inner\\[-2pt]entities};

\node[box, fill=gray!20, minimum width=1.0cm, below=0.4cm of inner2] (merge)
  {\scriptsize Merge};
\node[outputbox, right=0.2cm of merge] (final)
  {All entities\\[-2pt]{\scriptsize(outer+inner)}};

\draw[arr] (input2) -- (binder2);
\draw[arr] (binder2) -- (outer2);
\draw[arr] (outer2) -- (llm2);
\draw[arr] (llm2) -- (inner2);
\coordinate (outer-drop) at (outer2.south |- merge.west);
\draw[arr] (outer2.south) -- (outer-drop) -- (merge.west);
\draw[arr] (inner2) -- (merge);
\draw[arr] (merge) -- (final);

\end{tikzpicture}%
}
  \caption{Overview of methods for learning nested NER from flat annotations. (a)~Fine-tuned approaches augment flat data with pseudo-nested signal (inclusions, entity corruption) and modify training (flat neutralization) before training a Binder model. (b)~LLM-based approaches use either a pure few-shot LLM or a hybrid pipeline where a fine-tuned Binder detects outer entities and an LLM identifies inner entities within each span.}
  \label{fig:overview}
\end{figure*}
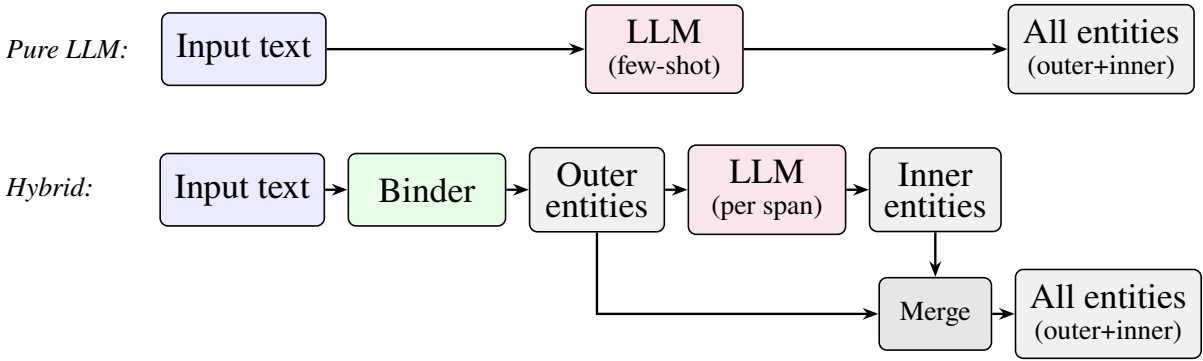

Our experiments reveal several findings. String inclusions recover substantial nested structure, improving inner entity F1 from 3.84\% to 21.36\%. Among entity corruption strategies, end-position corruption consistently outperforms alternatives. Combining all three fine-tuning methods---inclusions, corruption, and neutralization---achieves 26.37\% inner F1, closing 40\% of the gap to full supervision. The hybrid fine-tuned + LLM approach achieves 70.16\% overall F1 but underperforms fine-tuned methods on inner entities, indicating that current LLMs struggle with fine-grained nested structure across many entity types.

Our contributions are: (1) a systematic comparison of methods for learning nested NER from flat annotations (inclusions, corruption, neutralization) on a challenging 29-type Russian benchmark; (2) analysis of entity corruption strategies, finding that end-position corruption consistently outperforms alternatives; (3) a hybrid fine-tuned + LLM pipeline with evaluation of its limitations; (4) empirical evidence that simple methods close 40\% of the gap between flat and full supervision.

\section{Related work}

Multiple approaches address nested NER: span-based biaffine parsing \citep{yu-etal-2020-named}, layered models \citep{ju-etal-2018-neural,wang-etal-2020-pyramid}, set prediction \citep{tan-etal-2021-sequence}, hypergraph methods \citep{yan-etal-2023-local}, and bi-encoder approaches with contrastive learning \citep{zhang-etal-2023-binder} that represent entity types through natural language descriptions. All these approaches assume fully-annotated nested training data.

When annotations are incomplete, distant supervision \citep{lison-etal-2020-named,meng-etal-2021-distantly} and data augmentation \citep{dai-adel-2020-analysis} provide alternatives. Our inclusions and corruption methods can be viewed as augmentation strategies specifically designed for generating pseudo-nested training signal.

As for learning nested entities from flat annotations, \citet{zhu-etal-2022-recognizing} propose excluding within-entity spans from negative sampling during training, achieving 54.8\% nested F1 on ACE 2004. We build on this insight with a content-aware variant that selectively neutralizes spans matching known entity surface forms (Section~\ref{sec:neutralization}). \citet{rozhkov-loukachevitch-2025-methods} investigate this task on nested terms, demonstrating growing interest in learning nested structures from flat supervision.

LLM-based approaches \citep{wang-etal-2023-gptner} show prospect for NER but struggle with structured extraction tasks. \citet{han-etal-2023-information} demonstrate that ChatGPT significantly underperforms fine-tuned models on information extraction, with performance degrading further for complex structures and large type inventories. These limitations motivate hybrid approaches combining fine-tuned models with LLM reasoning.

Several corpora support nested NER research. Early benchmarks include ACE 2004/2005 \citep{doddington-etal-2004-automatic} with 7 English entity types and GENIA \citep{kim-etal-2003-genia} for biomedical English. NNE \citep{ringland-etal-2019-nne} extends to 114 fine-grained types. Recent work has expanded nested NER to diverse languages and domains: DaN+ for Danish \citep{plank-etal-2020-dan}, Wojood for Arabic \citep{jarrar-etal-2022-wojood}, historical documents \citep{tual-etal-2023-benchmark}, judicial Chinese \citep{zhang-etal-2023-judicial}, and BioNNE for Russian biomedical texts \citep{davydova-etal-2024-bionne}. For Russian general-domain text, NEREL \citep{loukachevitch-etal-2021-nerel} provides 56K nested annotations across 29 types---the largest type inventory among major nested NER benchmarks---with 21\% of entities nested. The RuNNE-2022 shared task \citep{artemova-etal-2022-runne} established evaluation benchmarks on NEREL.

Building on these foundations, we systematically compare methods for learning nested NER from flat annotations on this challenging 29-type benchmark, including content-aware neutralization and hybrid fine-tuned+LLM pipelines.

\section{Methodology}

Figure~\ref{fig:overview} presents an overview of all approaches investigated in this work. We explore four complementary methods for recovering nested entities from flat annotations, organized into three categories: data augmentation techniques that create pseudo-nested training signal (Section~\ref{sec:data-aug}), training modifications that adjust how the model learns from flat data (Section~\ref{sec:training-mod}), and LLM-based approaches that leverage large language models (Section~\ref{sec:llm-approaches}).

\subsection{Problem formulation}

Given a text sequence $x = (x_1, x_2, \ldots, x_n)$ of $n$ tokens, nested NER aims to identify all entity spans $e = (i, j, t)$ where $i$ and $j$ are the start and end positions, and $t$ is the entity type. In nested NER, multiple entities can share the same tokens, allowing for overlapping and hierarchical structures.

In contrast, flat NER requires that no two mentions overlap. When converting nested annotations to flat annotations, we retain only the outermost mentions, removing all nested mentions within them. This creates a dataset where $\forall e_1, e_2$: either $e_1 \cap e_2 = \emptyset$ or one entity completely contains the other (but only the outer one is annotated).

\subsection{Data augmentation}
\label{sec:data-aug}

We first describe two methods that augment flat training data with pseudo-nested annotations.

\subsubsection{Inclusions}
\label{sec:inclusions}

We add inclusions to the training data---subsequences of flat mentions that match the surface form of other mentions in the dataset.

Formally, for each flat mention $e_i = (s_i, t_i)$ with text span $s_i$ and type $t_i$, we extract all other flat mentions $\{e_j | j \neq i\}$ from the training set, find all strict substring matches (if $s_j$ appears as a substring within $s_i$ and $s_j \neq s_i$, create a new mention at that position with type $t_j$), and add these new mentions to the training data as positive examples.

For morphologically rich languages like Russian, the same entity may appear in different grammatical cases. We therefore also evaluate \textit{lemmatized inclusions}: each mention is tokenized and each token is reduced to its dictionary form using pymorphy2 \citep{korobov2015morphological}, producing a canonical representation (sorted lemmatized tokens). Substring matching is then performed on these canonical forms rather than raw surface strings, recovering additional inclusions across inflectional variants. Lemmatization is applied only during training data preparation; at test time the model predicts spans directly from raw text.

\subsubsection{Entity corruption}

Another approach to create nested annotations is through ``entity corruption''. The idea is to train the model to recognize that when a word within a long entity is corrupted (replaced with a noise token), the remaining parts might still be valid entities.

We experiment with two corruption strategies, \textit{early damage} and \textit{late damage}.

In \textit{early damage} strategy, we split the training data into 5 folds. For each fold, we train the model on 80\% of the data where long flat mentions (length $> 2$ words) have one word corrupted, predict on the remaining 20\% (uncorrupted), combine predictions from all folds to create pseudo-nested annotations, and retrain on the combination of flat and pseudo-nested data.

In \textit{late damage} strategy, similar to first, but we train the model on 80\% of uncorrupted data, then predict on the remaining 20\% with corrupted entities. The model learns to identify entities within corrupted contexts.

For illustration purposes, let us consider the flat entity ``Ministry of Foreign Affairs of Russia'' (ORGANIZATION). With end-position corruption using letters, the last word is replaced: ``Ministry of Foreign Affairs of \textit{klr}''. In Strategy~2, a model trained on clean data predicts on this corrupted sentence and may recognize ``Ministry of Foreign Affairs'' as ORGANIZATION---a pseudo-nested annotation. Across the corpus, such predictions are collected and used as additional training signal.

\textbf{Corruption symbols:} We explore five types of replacement symbols: digits (e.g., ``798'')---random digit sequences unlikely to form meaningful words; letters (e.g., ``klr'')---random consonant sequences that violate phonotactic rules; diglets (e.g., ``9z2'')---mixed digits and letters for maximum ``jumbledness''; semicolons (e.g., ``;;;'')---punctuation marks that signal phrase boundaries; and commas (e.g., ``,,,''). The motivation for alphanumeric corruption is that these sequences are too ``jumbled'' to be interpreted as real words, acting as neutral placeholders. Punctuation-based corruption was hypothesized to provide stronger boundary signals but also affects subword tokenization more aggressively.

\textbf{Corruption positions:} We experiment with five positions within multi-word entities: start (corrupt the first word), end (corrupt the last word), middle (corrupt the middle word), random (corrupt a randomly selected word), and syntax (corrupt the syntactic root identified using dependency parsing). Different positions test different hypotheses: corrupting the end position preserves the entity's head word (often the first word in Russian noun phrases), while corrupting the start tests whether the model can recognize entities from their modifiers alone.

\subsection{Fine-tuned model and training modifications}
\label{sec:training-mod}

\subsubsection{Binder model}
\label{sec:baseline}

Our baseline approach trains a nested NER model (Binder \citep{zhang-etal-2023-binder}) on flat annotations. Binder is a bi-encoder span-based model that achieves state-of-the-art results on several nested NER benchmarks including ACE 2004 and GENIA. It represents entity types through natural language prompts and predicts entities for all possible spans in a sentence using contrastive learning.

During training, Binder classifies each candidate span as either a positive entity match or a negative (non-entity) example. When training on flat data, the model learns to classify outer mention spans as positive and all other spans---including potential nested mentions---as negative.

The key challenge is that subsequences within flat mentions are treated as negative examples during training, even though they might be valid nested mentions in the true annotation. Our neutralization method (Section~\ref{sec:neutralization}) addresses this by introducing a third category: neutral spans that are neither positive nor negative during training.

\subsubsection{Binder prompts}

The Binder model uses natural language \textit{type descriptions} to represent entity types in its bi-encoder architecture (distinct from LLM prompting in Section~\ref{sec:llm-approaches}). We evaluate six description strategies, following \citep{rozhkov-2023}: keyword (entity type name only), definition (natural language definition), most-frequent-class (most common entity example per type), context (sentence contexts), lexical-all-outer (full sentences with entity markers), and struct-nested (structural nesting information). This verifies that our weak supervision methods generalize across different type description configurations. Full details are provided in Appendix~\ref{sec:full-prompt-comparison}.

\subsubsection{Flat neutralization}
\label{sec:neutralization}

\citet{zhu-etal-2022-recognizing} show that excluding within-entity spans from negative sampling improves nested entity recognition from flat supervision. Their approach uniformly ignores all spans geometrically contained within annotated entities. We propose a content-aware variant: rather than ignoring all within-entity spans, we selectively neutralize only those matching known entity surface forms via inclusion matching (Section~\ref{sec:inclusions}), retaining negative signal for non-matching spans.

Specifically, we partition candidate spans into three sets: positive $\mathcal{P}$ (annotated mentions), negative $\mathcal{N}$ (spans not matching any known entity), and neutral $\mathcal{U}$ (within-entity spans matching known entities via inclusion). The training loss excludes neutral spans:
\begin{equation}
\mathcal{L}_{\text{neutral}}(s) = \begin{cases}
\mathcal{L}(s) & \text{if } s \in \mathcal{P} \cup \mathcal{N} \\
0 & \text{if } s \in \mathcal{U}
\end{cases}
\end{equation}

This approach can be combined with inclusions: adding matched spans as positive examples while neutralizing remaining potential nested positions.

\subsection{LLM-based approaches}
\label{sec:llm-approaches}

\subsubsection{Pure LLM}

We investigate prompt-based approaches using DeepSeek-R1-32B \citep{deepseek-r1-2025} and RuAdapt-Qwen2.5-32B \citep{tikhomirov2024facilitating,tikhomirov2023impact}. We selected DeepSeek-R1-32B for its strong reasoning capabilities and RuAdapt-Qwen2.5-32B as a Russian-adapted model, both suitable for local deployment without API costs. Prompts explicitly define two nesting types \citep{kim-etal-2024-exploring}: NDT (different type nesting) and NST (same type hierarchy), with definitions for all 29 entity classes (Appendix~\ref{sec:appendix}). Output format is JSON: \texttt{\{entity : type\}}.

We compare zero-shot, one-shot, and five-shot configurations with three example selection strategies:

\textbf{Random sampling:} Examples selected uniformly at random from the training set.

\textbf{Most-frequent-entity (MFE):} Sentences containing the most frequently occurring entities, ensuring coverage of common types.

\textbf{Entity-wise selection (MFE-entwise):} For each of the 29 entity types, we select top examples by frequency, ensuring balanced coverage. The variant MFE-entwise-sent operates at sentence level with morphological normalization, grouping entity mentions by lemmatized forms to handle Russian inflection.

Figure~\ref{fig:selection-strategies} illustrates the key differences between strategies.

\begin{figure}[t]
\small
\textbf{Random:} Examples selected uniformly at random from the training set.\\[3pt]
\textbf{MFE:} Sentences containing the most frequently occurring entities, ensuring coverage of common types. Top entities: \textit{\foreignlanguage{russian}{США}~(USA)}, \textit{\foreignlanguage{russian}{России}~(Russia)}, \textit{\foreignlanguage{russian}{Москве}~(Moscow)}, \textit{\foreignlanguage{russian}{Владимир Путин}~(Vladimir Putin)}, \ldots\\[3pt]
\textbf{MFE-entwise:} For each of the 29 entity types, the top examples by frequency are selected, ensuring balanced coverage. The per-type list appended to the prompt:\\[2pt]
{\scriptsize
\begin{tabular}{@{}l@{~~}l@{}}
CITY: & \textit{\foreignlanguage{russian}{Москве, Нью-Йорке, Лондоне}} \\
 & \textit{(Moscow, New York, London)} \\
COUNTRY: & \textit{\foreignlanguage{russian}{США, России, Россия}} \\
 & \textit{(USA, of Russia, Russia)} \\
PERSON: & \textit{\foreignlanguage{russian}{Владимир Путин, Путин, Обама}} \\
 & \textit{(Vladimir Putin, Putin, Obama)} \\
\multicolumn{2}{@{}l@{}}{...~[all 29 types]}
\end{tabular}
}\\[3pt]
\textbf{MFE-entwise-sent:} Same as MFE-entwise, but operates at sentence level with morphological normalization, grouping entity mentions by lemmatized forms: \textit{\foreignlanguage{russian}{России/Россия/Россию}~(of Russia / Russia / Russia\textsubscript{acc})} $\rightarrow$ \textit{\foreignlanguage{russian}{Россия}~(Russia)}.
\caption{Example selection strategies for LLM prompts. MFE selects sentences by entity frequency; MFE-entwise adds a per-type entity list; MFE-entwise-sent additionally applies morphological normalization.}
\label{fig:selection-strategies}
\end{figure}

To improve few-shot effectiveness, we documented characteristic nesting patterns for each entity class (Appendix~\ref{sec:appendix2}).

\subsubsection{Hybrid fine-tuned + LLM}

We propose a two-stage hybrid pipeline. First, a fine-tuned Binder model processes the input text to detect outer entities (82--83\% F1 on this subtask). Second, for each detected outer entity span, we extract the text and query an LLM to identify nested entities within that span. The LLM receives the outer entity text, its predicted type, and a prompt describing common nesting patterns for that type (Appendix~\ref{sec:appendix2}). Predicted nested entities are merged with outer entities from stage one.

This decomposition leverages the strengths of both approaches: fine-tuned models excel at boundary detection and type classification, while LLMs can apply reasoning about entity composition. We test three variants: (a) \textit{type-specific}: prompts tailored to each outer entity type's nesting patterns, (b) \textit{full prompts}: comprehensive prompts covering all 29 types, and (c) \textit{lemmatization matching}: morphological normalization to handle Russian inflection.

Inclusions and entity corruption were introduced in our prior work on nested term extraction \citep{rozhkov-loukachevitch-2025-methods}, which we adapt here to the more challenging setting of 29-type named entity recognition. Flat neutralization builds on \citet{zhu-etal-2022-recognizing}'s insight of excluding within-entity spans from negative sampling, with our contribution being content-aware selection via inclusion matching. The hybrid fine-tuned+LLM pipeline is novel to this work. Our primary contribution is the systematic comparison and combination of these methods on a challenging benchmark, along with new analysis of corruption position effects.

\section{Experimental setup}

\subsection{Datasets}

NEREL is a Russian nested NER dataset containing 29 entity types including PERSON, ORGANIZATION, LOCATION, DATE, etc. The dataset consists of news texts with rich nested entity annotations: train contains 44,680 entities (35,340 outer, 9,340 inner), dev contains 5,541 entities (4,587 outer, 954 inner), and test contains 5,790 entities (4,745 outer, 1,045 inner). Nesting depth reaches up to 6 levels, though the vast majority of nesting is shallow: across all splits, 83.4\% of inner entities are at depth 2 (directly inside an outer entity), with only 3.5\% of all entities at depth 3 or beyond. The RuNNE-2022 shared task \citep{artemova-etal-2022-runne} evaluated nested NER on NEREL with a reduced training set; our full-supervision Binder baseline surpasses shared task results, so we use it as the upper bound. For flat training, we created the NEREL-outerflat dataset by removing all inner entities, retaining only the 35,340 outermost entities in the training set.

\subsection{Model}

We use Binder (described in Section~\ref{sec:baseline}) built on RuRoBERTa-large \citep{zmitrovich-etal-2024-family}. Training uses the AdamW optimizer \citep{loshchilov2019decoupled} with a learning rate of 1e-5, batch size of 8, and 64 epochs on a single NVIDIA RTX 4090 GPU. We run each experiment 5 times with different random seeds and report mean {\scriptsize$\pm$} standard deviation. LLM experiments use DeepSeek-R1-32B and RuAdapt-Qwen2.5-32B (both 32B parameters), served locally on modern accelerators using the vLLM framework (v0.8.3). Each LLM experiment is run with 3 seeds.

For LLM experiments, we use temperature of 0 for near-deterministic outputs, repetition penalty of 1.05--1.1, top-$p$ of 1.0, and maximum output length of 5,000 tokens. LLMs return a JSON dictionary \texttt{\{entity: type\}} enclosed in triple backticks. We extract text between backticks, strip formatting artifacts, and parse as JSON. Each predicted entity string is matched to the source text by exact string occurrence to determine character offsets; if a predicted string does not appear in the source text, it is discarded. Predicted types not in the 29-type inventory are retained but will not match any gold entity during evaluation. Failed JSON parses are treated as empty predictions (no entities for that input).

\subsection{Evaluation metrics}

Following standard practice in nested NER evaluation \citep{lu-roth-2015-joint,zhang-etal-2023-binder}, we report micro F1 and macro F1 scores computed separately for three entity categories.

For each predicted and gold entity, we classify it as inner or outer based on containment relationships within the same set (predictions or gold, respectively): an entity $e_i$ is inner if there exists another entity $e_j$ in the same set such that $e_j$ fully contains $e_i$ (i.e., $\text{start}(e_j) \leq \text{start}(e_i)$ and $\text{end}(e_i) \leq \text{end}(e_j)$), and outer otherwise. This means inner/outer classification is performed independently for gold and predicted entities. A predicted entity is compared against other predicted entities to determine if it is inner, and similarly for gold entities.

We compute overall F1 over all entities, inner F1 computed only on inner gold entities vs. inner predicted entities, and outer F1 computed only on outer gold entities vs. outer predicted entities. Micro F1 aggregates true positives, false positives, and false negatives across all documents before computing F1. Macro F1 computes F1 per document and averages.

\section{Results}

\subsection{Inclusion statistics}

For NEREL, we extracted 6,481 inclusions from the flat training data (18.34\% of the 35,340 flat mentions). The distribution across entity types is highly skewed, with PERSON (2,851), PROFESSION (888), ORGANIZATION (668), and COUNTRY (622) being the most frequent. Similar patterns appear in dev (739 inclusions, 16.71\%) and test (814 inclusions, 17.68\%) sets.

Named entities show high surface form reuse, with 18.34\% of flat mentions containing potential nested mentions identified through substring matching. The distribution across entity types varies substantially: COUNTRY and ORGANIZATION inclusions have 82.55\% and 59.82\% precision respectively, while PERSON inclusions---the most numerous---have only 2.81\% precision due to name components matching across unrelated entities (Appendix~\ref{sec:inclusion-per-type}).

Comparing the 6,458 exact inclusions against gold nested annotations, 1,488 match true inner entities (23.04\% precision, 17.48\% recall). When an inclusion span does match a gold inner entity, the type is correct 98.9\% of the time (1,488 of 1,504 span matches), which confirms that substring matching reliably assigns the correct type. The remaining 76.7\% of inclusions are spurious---their spans do not correspond to any gold inner entity. Lemmatized inclusions produce far more candidates (120,420) but with much lower precision (0.51\%) and recall (7.22\%). Despite this noise, training with inclusions substantially improves inner entity detection (3.84\% $\rightarrow$ 21.36\% F1), which suggests that even approximate nested signal provides some useful supervision.

\subsection{Flat vs. inclusions vs. full training}

Table~\ref{tab:main-results} presents our main experimental results comparing flat training, training with inclusions, and full nested training on NEREL. We show results for the two best-performing prompt strategies (Keyword and Lex-all-outer); full comparison across all six prompt types is provided in Appendix~\ref{sec:full-prompt-comparison}.

\begin{table}[t]
\centering
\small
\begin{tabular}{llccc}
\toprule
\textbf{Training} & \textbf{Prompt} & \textbf{Ovrl.} & \textbf{Inner} & \textbf{Outer} \\
\midrule
Flat & Keyword & 76.59 & 3.84 & 83.28 \\
Flat & Lex-all-out & 76.72 & 3.11 & 83.40 \\
\midrule
Flat+Incl. & Keyword & 76.70 & 21.36 & 83.05 \\
Flat+Incl. & Lex-all-out & 76.93 & 22.16 & 83.17 \\
\midrule
Full nested & Keyword & \textbf{85.81} & \textbf{65.48} & \textbf{83.95} \\
Full nested & Lex-all-out & 85.55 & 65.33 & 83.59 \\
\bottomrule
\end{tabular}
\caption{Baseline comparison (Micro F1, \%) on NEREL test set. Flat and Flat+Inclusions represent weak supervision; Full nested is the upper bound with complete annotations. Full prompt comparison in Appendix~\ref{sec:full-prompt-comparison}.}
\label{tab:main-results}
\end{table}

Flat training yields only 3--4\% inner F1, demonstrating that models trained solely on outer entities cannot recognize nested structures. Adding inclusions increases inner F1 to 21--22\%---a 5--7x improvement---with minimal degradation in outer entity performance. However, full nested training achieves 65\% inner F1, indicating that inclusion-based methods, while effective, cannot replace full supervision. Different prompts yield similar results within each training regime (differences within 1--2\% F1).

\subsection{Entity corruption results}

Table~\ref{tab:corruption-summary} summarizes the best corruption strategies for each symbol type. Full results across all positions and strategies are provided in Appendix~\ref{sec:full-corruption-results}.

\begin{table}[t]
\centering
\small
\begin{tabular}{llccc}
\toprule
\textbf{Symbol} & \textbf{Position} & \textbf{Overall} & \textbf{Inner} & \textbf{Outer} \\
\midrule
Digits & end & 77.85 & 23.96 & \textbf{82.61} \\
Letters & end & 77.81 & \textbf{25.92} & 82.39 \\
Diglets & start & \textbf{78.08} & 22.43 & 82.27 \\
Semicolon & start & 74.59 & 19.16 & 77.13 \\
Comma & start & 74.23 & 19.46 & 76.38 \\
\bottomrule
\end{tabular}
\caption{Best entity corruption results (Micro F1, \%) for each symbol type using Strategy 1 (train on corrupted, predict on clean). Full results in Appendix~\ref{sec:full-corruption-results}.}
\label{tab:corruption-summary}
\end{table}

Training on corrupted data (Strategy 1) consistently outperforms corrupting at test time (Strategy 2). End-position corruption yields the highest inner F1 (25.92\% with letters), likely because entity beginnings carry more type information. Alphanumeric corruption (digits, letters, diglets) maintains strong overall F1 (77--78\%) while achieving 22--26\% inner F1. Punctuation-based corruption (semicolons, commas) substantially degrades performance.

\subsection{Flat neutralization results}

Table~\ref{tab:neutralization-results} presents results for flat neutralization approaches, which mark potential nested mention positions as neutral during training rather than treating them as negative examples.

\begin{table}[t]
\centering
\small
\resizebox{\columnwidth}{!}{%
\begin{tabular}{lccc}
\toprule
\multirow{2}{*}{\textbf{Approach}} & \multicolumn{3}{c}{\textbf{Micro F1 (\%)}} \\
\cmidrule(lr){2-4}
& Overall & Inner & Outer \\
\midrule
\multicolumn{4}{l}{\textit{Baseline}} \\
Flat & 76.59{\scriptsize$\pm$0.23} & 3.84{\scriptsize$\pm$0.77} & \textbf{83.28}{\scriptsize$\pm$0.23} \\
Inclusions & 76.70{\scriptsize$\pm$0.17} & 21.36{\scriptsize$\pm$1.53} & 83.05{\scriptsize$\pm$0.26} \\
\midrule
\multicolumn{4}{l}{\textit{Neutralization}} \\
Flat+Neutral. & 76.82{\scriptsize$\pm$0.19} & 5.43{\scriptsize$\pm$0.81} & 83.01{\scriptsize$\pm$0.22} \\
Incl.+Neutral. & 77.01{\scriptsize$\pm$0.15} & 22.68{\scriptsize$\pm$1.27} & 82.93{\scriptsize$\pm$0.18} \\
Lem.Incl.+Neutral. & 77.23{\scriptsize$\pm$0.21} & 24.15{\scriptsize$\pm$1.02} & 82.78{\scriptsize$\pm$0.26} \\
\midrule
\multicolumn{4}{l}{\textit{+Corruption}} \\
Lem.Incl.+Corr.+Neu. & \textbf{77.89}{\scriptsize$\pm$0.18} & \textbf{26.37}{\scriptsize$\pm$0.94} & 82.54{\scriptsize$\pm$0.21} \\
\bottomrule
\end{tabular}%
}
\caption{Results for flat neutralization approaches on NEREL test set. Neutralization marks potential nested mentions as neutral (neither positive nor negative) during training. ``Lem.Incl.'' refers to lemmatized inclusions.}
\label{tab:neutralization-results}
\end{table}

Flat neutralization alone improves inner F1 from 3.84\% to 5.43\%---a 41\% relative improvement without explicit positive examples. Combining neutralization with inclusions (22.68\%) modestly outperforms inclusions alone (21.36\%). Lemmatization further improves matching by handling morphological variants (24.15\%). Our best combined method (lemmatized inclusions + corruption + neutralization) achieves 26.37\% inner F1 and 77.89\% overall F1, closing 40\% of the gap to full supervision while maintaining strong outer entity performance (82.54\%).

\subsection{LLM results}

Table~\ref{tab:llm-results} presents LLM and hybrid results. Pure LLM approaches substantially underperform fine-tuned methods: DeepSeek-R1 achieves only 42.39\% overall F1 (5-shot) compared to 76.59\% for flat training. The hybrid approach (70.16\% overall F1) outperforms pure LLM methods but still underperforms fine-tuned approaches on inner entities (18.84\% vs 21.36\%). This suggests current LLMs struggle with fine-grained nested structure across 29 entity types.

\begin{table}[t]
\centering
\small
\resizebox{\columnwidth}{!}{%
\begin{tabular}{lccc}
\toprule
\textbf{Method} & \textbf{Overall} & \textbf{Inner} & \textbf{Outer} \\
\midrule
\multicolumn{4}{l}{\textit{Traditional Approaches}} \\
Flat & 76.59{\scriptsize$\pm$0.23} & 3.84{\scriptsize$\pm$0.77} & \textbf{83.28}{\scriptsize$\pm$0.23} \\
Flat+Inclusions & \textbf{76.70}{\scriptsize$\pm$0.17} & \textbf{21.36}{\scriptsize$\pm$1.53} & 83.05{\scriptsize$\pm$0.26} \\
\midrule
\multicolumn{4}{l}{\textit{LLM Approaches (MFE)}} \\
DeepSeek-R1 (0-shot) & 38.35 & 2.69 & 38.86 \\
DeepSeek-R1 (1-shot) & 39.15 & 4.57 & 40.14 \\
DeepSeek-R1 (5-shot) & \textbf{42.39} & \textbf{5.78} & \textbf{42.63} \\
RuAdapt (0-shot) & 29.98 & 1.77 & 32.68 \\
RuAdapt (1-shot) & 29.23 & 2.00 & 31.82 \\
RuAdapt (5-shot) & 31.89 & 3.91 & 34.02 \\
\midrule
RuAdapt (MFE-entwise-sent) & 45.91 & 2.42 & 46.51 \\
\midrule
\multicolumn{4}{l}{\textit{Hybrid Approaches}} \\
Hybrid (type-specific) & 66.92 & \textbf{20.24} & \textbf{82.73} \\
Hybrid (full prompts) & 69.28 & 17.40 & 82.73 \\
Hybrid (lem. matching) & \textbf{70.16} & 18.84 & 75.83 \\
\bottomrule
\end{tabular}%
}
\caption{LLM and hybrid results on NEREL test set (Micro F1, \%). Traditional approaches show mean{\scriptsize$\pm$}std across 3 runs; LLM approaches show mean across 3 seeds.}
\label{tab:llm-results}
\end{table}

\subsection{Summary comparison}

Table~\ref{tab:summary} presents a unified comparison of the best-performing variant from each method category.

\begin{table}[t]
\centering
\small
\setlength{\tabcolsep}{4pt}
\begin{tabular}{lccc}
\toprule
\textbf{Method} & \textbf{Overall} & \textbf{Inner} & \textbf{Outer} \\
\midrule
\multicolumn{4}{l}{\textit{Weak Supervision (Fine-tuned)}} \\
Flat (baseline) & 76.59 & 3.84 & 83.28 \\
+ Inclusions & 76.70 & 21.36 & 83.05 \\
+ Corruption (letters, end) & 77.81 & 25.92 & 82.39 \\
+ Lem.Incl.+Corr.+Neutral. & \textbf{77.89} & \textbf{26.37} & \textbf{82.54} \\
\midrule
\multicolumn{4}{l}{\textit{LLM-based}} \\
Pure LLM (DeepSeek 5-shot) & 42.39 & 5.78 & 42.63 \\
Hybrid (lem. matching) & 70.16 & 18.84 & 75.83 \\
\midrule
\multicolumn{4}{l}{\textit{Upper Bound}} \\
Full nested supervision & \underline{85.81} & \underline{65.48} & \underline{83.95} \\
\bottomrule
\end{tabular}
\caption{Summary comparison of best methods (Micro F1, \%). Our best weak supervision method (Lem.Incl.+Corr.+Neutral.) closes 40\% of the inner F1 gap between flat training and full supervision.}
\label{tab:summary}
\end{table}

The comparison reveals a clear hierarchy: pure LLM approaches substantially underperform fine-tuned methods, while hybrid approaches fall between pure LLM and fine-tuned weak supervision. Our best combined method achieves the highest inner F1 among weak supervision approaches, closing 40\% of the gap to full nested supervision.

\section{Conclusion}

We investigated methods for learning nested named entity recognition from flat annotations, addressing the annotation cost that limits nested NER development. Using the Russian NEREL dataset with 29 entity types, we systematically compared string inclusions, entity corruption, flat neutralization, and hybrid fine-tuned with LLM approaches.

Our experiments demonstrate that nested structure can be recovered without gold nested annotations. String inclusions alone improve inner entity F1 from 3.84\% to 21.36\% by leveraging substring relationships between entities. Entity corruption creates pseudo-nested training examples, with our analysis revealing that end-position corruption consistently outperforms other positions---a finding not previously documented. Combining these methods with flat neutralization achieves 26.37\% inner F1, closing 40\% of the gap to full nested supervision.

However, a significant performance gap remains. Full nested supervision achieves 65.48\% inner F1, indicating that our weak supervision methods, while effective, cannot fully substitute for nested annotations. The hybrid fine-tuned+LLM approach, despite leveraging large language model capabilities, underperforms fine-tuned methods on inner entity detection, suggesting that current LLMs struggle with fine-grained nested structure across many entity types.

Returning to our research questions:

\textbf{RQ1:} Yes, models can learn to recognize nested entities from flat annotations. String inclusions alone achieve 21.36\% inner F1, and combining methods reaches 26.37\%, closing 40\% of the gap to full supervision.

\textbf{RQ2:} Partially. The hybrid fine-tuned + LLM pipeline achieves 70.16\% overall F1, leveraging fine-tuned models for reliable outer entity detection (82--83\% F1) while using LLMs for nested structure. However, this approach underperforms purely fine-tuned methods on inner entities (18.84\% vs 26.37\% inner F1), indicating that current LLMs provide limited benefit for fine-grained nested detection across many entity types.

\textbf{RQ3:} No, current LLMs cannot compensate for missing nested annotations. Pure LLM approaches achieve only 42.39\% overall F1 compared to 76.59\% for fine-tuned models, with inner entity detection at just 5.78\% F1. LLMs struggle with the 29-type inventory, frequently hallucinating entity boundaries and types. While LLMs show some prospect for coarse-grained NER, they cannot substitute for proper supervision when fine-grained nested structure is required.

Future work should investigate more principled approaches to negative sampling in span-based models, cross-lingual transfer of weak supervision methods, and improved LLM prompting strategies for nested structure. The gap between weak and full supervision also motivates research into annotation-efficient approaches that selectively annotate nested structure where it provides the most benefit.

\section*{Limitations}

Our experiments focus exclusively on Russian, using the NEREL dataset. While Russian's rich morphology provides a challenging test case, our findings may not generalize to other languages---especially those with very different structural properties. Morphological normalization is inherently language-specific, and its effectiveness in other languages remains to be verified. Additionally, because NEREL consists of news text, we cannot claim our results would hold in domains such as biomedical or social media data.

We evaluated on NEREL only, without cross-dataset validation. Although it is one of the largest nested NER resources available, testing on additional benchmarks would strengthen our conclusions. We leave experiments on other Russian nested NER datasets, such as NEREL-BIO, for future work.

All our comparisons use the Binder architecture; we did not test alternatives such as biaffine parsers or sequence-to-sequence models. Architecture comparisons are complicated by differences in preprocessing and evaluation, so we kept the focus narrow: our goal was to compare weak supervision strategies, not model designs. The strategies we studied (inclusions, corruption, neutralization) are architecture-agnostic and could be applied to other span-based NER models.

Class imbalance in NEREL is pronounced---PERSON and ORGANIZATION dominate. We did not adjust for this with re-sampling or loss weighting, so performance on rare types may be understated.

For our LLM experiments, we used DeepSeek-R1 and RuAdapt-Qwen2.5, chosen for strong Russian support and local deployability. Larger models such as GPT-4 may achieve better performance, but cost and reproducibility concerns ruled them out. Our LLM results should therefore be interpreted as demonstrating the approach rather than establishing upper bounds.

Results are averaged over three runs with standard deviations reported. We did not perform formal significance tests. With only a few seeds, smaller differences between methods should be treated cautiously, though the larger improvements (e.g., 3.84\% to 21.36\% inner F1) clearly exceed random variation.

Finally, we evaluated all experimental configurations on the test set rather than reserving it for final evaluation only. While we fixed hyperparameters upfront and our methods were linguistically motivated---not tuned on test metrics---this approach still carries some overfitting risk. A stricter design with dedicated development-phase tuning would have been stronger.

\section*{Acknowledgments}

The study was supported by the Russian Science Foundation project 25-21-00206. The research was carried out using the MSU-270 supercomputer of Lomonosov Moscow State University.

\bibliography{anthology,custom}

\appendix
\section{Base LLM prompt template}
\label{sec:appendix}

The base prompt template used for all pure LLM experiments:

\begin{lstlisting}
Given entity label set: ['AGE', 'AWARD', 'CITY', 'COUNTRY', 'CRIME', 'DATE', 'DISEASE', 'DISTRICT', 'EVENT', 'FACILITY', 'FAMILY', 'IDEOLOGY', 'LANGUAGE', 'LAW', 'LOCATION', 'MONEY', 'NATIONALITY', 'NUMBER', 'ORDINAL', 'ORGANIZATION', 'PENALTY', 'PERCENT', 'PERSON', 'PRODUCT', 'PROFESSION', 'RELIGION', 'STATE_OR_PROVINCE', 'TIME', 'WORK_OF_ART'].
You are an excellent linguist and annotator. Based on the given entity label set, please recognize the named entities in the given text. Consider there might be a nested case, where one entity contains another. There are two possible type of nested entities:
NDT: It consists of an entity containing a shorter entity tagged with a different type.
NST: This case usually occurs when entities are originally represented by a hierarchy.
Give me ONLY entities in format of a json dictionary with named entities as keys and their types as values like this: {entity : type}. Do not write any additional text. Enclose answer in ```.
\end{lstlisting}

For methods with entity definitions (e.g., type-specific hybrid), Russian definitions are appended after the base prompt.

\section{Entity-specific nesting patterns}
\label{sec:appendix2}

For hybrid fine-tuned+LLM approaches, we augment the base prompt with entity-specific nesting patterns derived from the training data. Below are examples for selected entity types:

\begin{Verbatim}[fontsize=\small, breaklines=true]
For class ORGANIZATION most common nested entity classes are: ORGANIZATION, COUNTRY, EVENT, CITY, PROFESSION

Here are some examples of the ORGANIZATION class as outermost entity and its nested entities:
Outermost entity: ```Федеральный штаб народного ополчения``` (Federal Headquarters of the People's Militia), nested are: ```[{"Федеральный штаб": "ORGANIZATION"}, (Federal Headquarters) {"штаб народного ополчения": "ORGANIZATION"}, (headquarters of the people's militia) {"ополчения": "ORGANIZATION"}]``` (militia)

For class PERSON most common nested entity classes are: PERSON, PROFESSION, ORDINAL, CITY, ORGANIZATION

Here are some examples of the PERSON class as outermost entity and its nested entities:
Outermost entity: ```Эрнст Теодор Амадей Гофман``` (Ernst Theodor Amadeus Hoffmann), nested are: ```[{"Эрнст": "PERSON"}, {"Амадей": "PERSON"}, {"Гофман": "PERSON"}]``` (Ernst, Amadeus, Hoffmann - name components)

For class DATE most common nested entity classes are: DATE, NUMBER, AGE, ORDINAL, PERSON

Here are some examples of the DATE class as outermost entity and its nested entities:
Outermost entity: ```21 октября 1952 года``` (October 21, 1952), nested are: ```[{"1952": "DATE"}, {"21": "ORDINAL"}, {"года": "DATE"}, {"октября": "DATE"}]``` (1952, 21, year, October)

... [patterns for all 29 entity classes]
\end{Verbatim}

\section{Russian entity definitions}
\label{sec:appendix3}

For experiments requiring entity definitions, we provide Russian descriptions with English translations:

\small
\begin{description}[leftmargin=0pt, labelindent=0pt, itemsep=2pt]
\item[AGE] \foreignlanguage{russian}{Возраст: Это число, которое показывает, сколько лет кому-то или чему-то.} --- A number that shows how old someone or something is.
\item[AWARD] \foreignlanguage{russian}{Награда: Это признание заслуг или достижений.} --- Recognition of merits or achievements.
\item[CITY] \foreignlanguage{russian}{Город: Место, где живут люди, обычно больше, чем деревня.} --- A place where people live, usually larger than a village.
\item[COUNTRY] \foreignlanguage{russian}{Страна: Большая территория с определенным населением и правительством.} --- A large territory with a defined population and government.
\item[CRIME] \foreignlanguage{russian}{Преступление: Действия, запрещенные законом.} --- Actions prohibited by law.
\item[DATE] \foreignlanguage{russian}{Дата: Указание времени, когда что-то произошло.} --- An indication of when something happened.
\item[DISEASE] \foreignlanguage{russian}{Болезнь: Состояние, при котором организм не работает нормально.} --- A condition in which the body does not function normally.
\item[DISTRICT] \foreignlanguage{russian}{Район: Часть страны или города, имеющая свои границы и управление.} --- Part of a country or city with its own boundaries and governance.
\item[EVENT] \foreignlanguage{russian}{Событие: Что-то важное, что произошло.} --- Something important that happened.
\item[FACILITY] \foreignlanguage{russian}{Объект инфраструктуры: Строение или место, используемое для определенной цели.} --- A building or place used for a specific purpose.
\item[FAMILY] \foreignlanguage{russian}{Семья: Группа людей, связанных кровными узами.} --- A group of people related by blood ties.
\item[IDEOLOGY] \foreignlanguage{russian}{Идеология: Набор идей и убеждений, определяющих поведение и политику.} --- A set of ideas and beliefs that determine behavior and policy.
\item[LANGUAGE] \foreignlanguage{russian}{Язык: Система общения, состоящая из слов и правил.} --- A communication system consisting of words and rules.
\item[LAW] \foreignlanguage{russian}{Закон: Правила, установленные государством.} --- Rules established by the state.
\item[LOCATION] \foreignlanguage{russian}{Место: Где что-то находится.} --- Where something is located.
\item[MONEY] \foreignlanguage{russian}{Деньги: Средства обмена, используемые для покупки товаров и услуг.} --- A medium of exchange used to purchase goods and services.
\item[NATIONALITY] \foreignlanguage{russian}{Национальность: Принадлежность к определенной стране или народу.} --- Belonging to a particular country or nation.
\item[NUMBER] \foreignlanguage{russian}{Число: Цифра или количество чего-либо.} --- A digit or quantity of something.
\item[ORDINAL] \foreignlanguage{russian}{Порядковый номер: Указывает на позицию в ряду.} --- Indicates a position in a sequence.
\item[ORGANIZATION] \foreignlanguage{russian}{Организация: Группировка людей с общей целью.} --- A group of people with a common purpose.
\item[PENALTY] \foreignlanguage{russian}{Наказание: Последствия за нарушение закона.} --- Consequences for violating the law.
\item[PERCENT] \foreignlanguage{russian}{Процент: Доля от целого, выраженная в сотых долях.} --- A fraction of the whole expressed in hundredths.
\item[PERSON] \foreignlanguage{russian}{Человек: Индивидуальное лицо.} --- An individual.
\item[PRODUCT] \foreignlanguage{russian}{Продукт: То, что создано или произведено для продажи или использования.} --- Something created or produced for sale or use.
\item[PROFESSION] \foreignlanguage{russian}{Профессия: Вид деятельности, которым человек зарабатывает на жизнь.} --- A type of activity by which a person earns a living.
\item[RELIGION] \foreignlanguage{russian}{Религия: Вера и система верований.} --- Faith and a system of beliefs.
\item[STATE\_OR\_PROVINCE] \foreignlanguage{russian}{Штат или провинция: Административная единица внутри страны.} --- An administrative unit within a country.
\item[TIME] \foreignlanguage{russian}{Время: Конкретный момент или период.} --- A specific moment or period.
\item[WORK\_OF\_ART] \foreignlanguage{russian}{Художественное произведение: Произведения искусства, созданные человеком.} --- Works of art created by humans.
\end{description}
\normalsize

\section{Inclusion statistics per entity type}
\label{sec:inclusion-per-type}

Table~\ref{tab:inclusion-per-type} shows inclusion counts and precision for each of the 29 entity types. \textit{Inclusions} are pseudo-nested entities identified by exact substring matching; \textit{Lem.\ inclusions} additionally apply morphological normalization. \textit{Precision} indicates the percentage of inclusions that correspond to true nested annotations.

\begin{table}[t]
\centering
\scriptsize
\setlength{\tabcolsep}{3pt}
\begin{tabular}{lrrrrr}
\toprule
\textbf{Type} & \textbf{True} & \textbf{Incl.} & \textbf{Prec.} & \textbf{Lem.} & \textbf{Prec.} \\
 & \textbf{inner} & & \textbf{(\%)} & \textbf{incl.} & \textbf{(\%)} \\
\midrule
PERSON & 433 & 2,845 & 2.81 & 41,347 & 0.07 \\
PROFESSION & 1,820 & 884 & 28.05 & 14,042 & 0.97 \\
ORGANIZATION & 2,001 & 662 & 59.82 & 13,456 & 1.43 \\
COUNTRY & 1,773 & 619 & 82.55 & 13,874 & 1.28 \\
NUMBER & 194 & 456 & 3.29 & 2,477 & 0.04 \\
EVENT & 424 & 300 & 25.00 & 14,059 & 0.09 \\
CITY & 634 & 159 & 37.11 & 4,186 & 0.74 \\
DATE & 325 & 111 & 9.91 & 3,813 & 0.18 \\
STATE\_OR\_PROV. & 213 & 73 & 72.60 & 807 & 1.49 \\
ORDINAL & 324 & 60 & 11.67 & 1,377 & 0.29 \\
PRODUCT & 30 & 53 & 20.75 & 1,203 & 0.42 \\
AWARD & 0 & 31 & 0.00 & 739 & 0.00 \\
IDEOLOGY & 0 & 31 & 0.00 & 1,030 & 0.00 \\
FACILITY & 53 & 23 & 56.52 & 1,453 & 0.21 \\
LAW & 0 & 21 & 0.00 & 763 & 0.00 \\
NATIONALITY & 52 & 21 & 9.52 & 760 & 0.00 \\
AGE & 17 & 18 & 0.00 & 867 & 0.00 \\
LOCATION & 127 & 17 & 29.41 & 469 & 0.64 \\
DISEASE & 0 & 16 & 0.00 & 260 & 0.00 \\
CRIME & 0 & 13 & 0.00 & 1,096 & 0.00 \\
WORK\_OF\_ART & 0 & 11 & 0.00 & 789 & 0.00 \\
LANGUAGE & 0 & 10 & 0.00 & 87 & 0.00 \\
PENALTY & 0 & 9 & 0.00 & 320 & 0.00 \\
RELIGION & 0 & 8 & 0.00 & 67 & 0.00 \\
DISTRICT & 59 & 5 & 40.00 & 74 & 1.35 \\
MONEY & 22 & 1 & 0.00 & 155 & 0.00 \\
TIME & 9 & 1 & 0.00 & 202 & 0.00 \\
FAMILY & 3 & 0 & 0.00 & 561 & 0.18 \\
PERCENT & 0 & 0 & 0.00 & 87 & 0.00 \\
\midrule
\textbf{Total} & \textbf{8,513} & \textbf{6,458} & \textbf{23.04} & \textbf{120,420} & \textbf{0.51} \\
\bottomrule
\end{tabular}
\caption{Inclusion statistics per entity type on training data, sorted by exact inclusion count. \textit{True inner}: gold nested entities. \textit{Incl.}: exact substring inclusions. \textit{Lem.\ incl.}: lemmatized substring inclusions. Precision measures the fraction of inclusions matching gold inner entities. Entity types like COUNTRY and ORGANIZATION have high exact-match precision, while PERSON produces many inclusions with low precision due to name components matching across unrelated entities. Lemmatization vastly increases inclusion counts but reduces precision, as morphological normalization over-generalizes matching.}
\label{tab:inclusion-per-type}
\end{table}

\section{Full prompt comparison}
\label{sec:full-prompt-comparison}

Table~\ref{tab:full-prompt-comparison} presents complete results across all six prompt strategies.

\begin{table*}[t]
\centering
\small
\begin{tabular}{llcccccc}
\toprule
\multirow{2}{*}{\textbf{Training}} & \multirow{2}{*}{\textbf{Prompt}} & \multicolumn{3}{c}{\textbf{Micro F1 (\%)}} & \multicolumn{3}{c}{\textbf{Macro F1 (\%)}} \\
\cmidrule(lr){3-5} \cmidrule(lr){6-8}
& & Overall & Inner & Outer & Overall & Inner & Outer \\
\midrule
\multicolumn{8}{l}{\textit{Flat Training (NEREL-outerflat)}} \\
& Keyword & 76.59{\scriptsize$\pm$0.23} & 3.84{\scriptsize$\pm$0.77} & 83.28{\scriptsize$\pm$0.23} & 76.67{\scriptsize$\pm$0.17} & 3.30{\scriptsize$\pm$0.58} & 82.94{\scriptsize$\pm$0.27} \\
& Definition & 76.43{\scriptsize$\pm$0.17} & 4.42{\scriptsize$\pm$0.52} & 82.96{\scriptsize$\pm$0.31} & 76.57{\scriptsize$\pm$0.17} & 3.76{\scriptsize$\pm$0.79} & 82.66{\scriptsize$\pm$0.32} \\
& Most-freq & 76.60{\scriptsize$\pm$0.21} & 3.66{\scriptsize$\pm$0.46} & 83.08{\scriptsize$\pm$0.56} & 76.67{\scriptsize$\pm$0.24} & 3.23{\scriptsize$\pm$0.53} & 82.80{\scriptsize$\pm$0.55} \\
& Context & 75.32{\scriptsize$\pm$0.17} & 3.32{\scriptsize$\pm$0.72} & 81.38{\scriptsize$\pm$0.23} & 75.42{\scriptsize$\pm$0.12} & 2.79{\scriptsize$\pm$0.69} & 81.05{\scriptsize$\pm$0.30} \\
& Lex-all-out & 76.72{\scriptsize$\pm$0.17} & 3.11{\scriptsize$\pm$0.61} & 83.40{\scriptsize$\pm$0.26} & 76.69{\scriptsize$\pm$0.18} & 2.76{\scriptsize$\pm$0.26} & 83.08{\scriptsize$\pm$0.20} \\
& Struct-nest & 76.57{\scriptsize$\pm$0.15} & 4.04{\scriptsize$\pm$1.15} & 82.88{\scriptsize$\pm$0.46} & 76.63{\scriptsize$\pm$0.15} & 3.39{\scriptsize$\pm$1.04} & 82.52{\scriptsize$\pm$0.39} \\
\midrule
\multicolumn{8}{l}{\textit{Flat Training with Inclusions}} \\
& Keyword & 76.70{\scriptsize$\pm$0.17} & 21.36{\scriptsize$\pm$1.53} & 83.05{\scriptsize$\pm$0.26} & 76.53{\scriptsize$\pm$0.24} & 18.17{\scriptsize$\pm$1.14} & 82.65{\scriptsize$\pm$0.29} \\
& Definition & 76.79{\scriptsize$\pm$0.29} & 21.97{\scriptsize$\pm$0.67} & 83.01{\scriptsize$\pm$0.26} & 76.66{\scriptsize$\pm$0.33} & 18.47{\scriptsize$\pm$0.60} & 82.69{\scriptsize$\pm$0.38} \\
& Most-freq & 76.89{\scriptsize$\pm$0.10} & 21.30{\scriptsize$\pm$0.85} & 82.88{\scriptsize$\pm$0.90} & 76.74{\scriptsize$\pm$0.09} & 17.95{\scriptsize$\pm$0.59} & 82.65{\scriptsize$\pm$0.91} \\
& Context & 75.39{\scriptsize$\pm$0.39} & 18.64{\scriptsize$\pm$1.77} & 81.04{\scriptsize$\pm$0.31} & 75.32{\scriptsize$\pm$0.38} & 15.76{\scriptsize$\pm$1.34} & 80.74{\scriptsize$\pm$0.39} \\
& Lex-all-out & 76.93{\scriptsize$\pm$0.19} & 22.16{\scriptsize$\pm$1.93} & 83.17{\scriptsize$\pm$0.27} & 76.82{\scriptsize$\pm$0.15} & 18.60{\scriptsize$\pm$1.45} & 82.90{\scriptsize$\pm$0.29} \\
& Struct-nest & 76.72{\scriptsize$\pm$0.09} & 20.50{\scriptsize$\pm$0.77} & 83.15{\scriptsize$\pm$0.25} & 76.61{\scriptsize$\pm$0.08} & 17.56{\scriptsize$\pm$0.73} & 82.87{\scriptsize$\pm$0.24} \\
\midrule
\multicolumn{8}{l}{\textit{Full Nested Training}} \\
& Keyword & 85.81{\scriptsize$\pm$0.19} & 65.48{\scriptsize$\pm$0.94} & 83.95{\scriptsize$\pm$0.28} & 85.60{\scriptsize$\pm$0.21} & 54.53{\scriptsize$\pm$0.70} & 83.56{\scriptsize$\pm$0.31} \\
& Definition & 85.82{\scriptsize$\pm$0.14} & 65.90{\scriptsize$\pm$0.79} & 83.89{\scriptsize$\pm$0.23} & 85.58{\scriptsize$\pm$0.19} & 54.93{\scriptsize$\pm$0.44} & 83.52{\scriptsize$\pm$0.26} \\
& Most-freq & 85.72{\scriptsize$\pm$0.21} & 65.68{\scriptsize$\pm$1.49} & 83.88{\scriptsize$\pm$0.22} & 85.45{\scriptsize$\pm$0.25} & 54.87{\scriptsize$\pm$1.83} & 83.41{\scriptsize$\pm$0.34} \\
& Context & 83.91{\scriptsize$\pm$0.07} & 54.62{\scriptsize$\pm$0.95} & 81.48{\scriptsize$\pm$0.09} & 83.69{\scriptsize$\pm$0.05} & 45.94{\scriptsize$\pm$0.86} & 80.99{\scriptsize$\pm$0.15} \\
& Lex-all-out & 85.55{\scriptsize$\pm$0.10} & 65.33{\scriptsize$\pm$0.62} & 83.59{\scriptsize$\pm$0.14} & 85.24{\scriptsize$\pm$0.16} & 54.26{\scriptsize$\pm$0.59} & 83.17{\scriptsize$\pm$0.22} \\
& Struct-nest & 85.68{\scriptsize$\pm$0.08} & 65.93{\scriptsize$\pm$0.39} & 83.85{\scriptsize$\pm$0.11} & 85.42{\scriptsize$\pm$0.16} & 54.92{\scriptsize$\pm$0.29} & 83.44{\scriptsize$\pm$0.13} \\
\bottomrule
\end{tabular}
\caption{Full prompt comparison on NEREL test set. Six prompt strategies tested: Keyword (entity type names), Definition (natural language definitions), Most-freq (most frequent entity example per type), Context (sentence contexts), Lex-all-out (full sentences with entity markers), Struct-nest (structural nesting information).}
\label{tab:full-prompt-comparison}
\end{table*}

\section{Full entity corruption results}
\label{sec:full-corruption-results}

Table~\ref{tab:full-corruption-results} presents complete results for all entity corruption strategies.

\begin{table*}[t]
\centering
\small
\begin{tabular}{llccccccc}
\toprule
\multirow{2}{*}{\textbf{Corruption}} & \multirow{2}{*}{\textbf{Position}} & \multicolumn{3}{c}{\textbf{Micro F1 (\%)}} & \multicolumn{3}{c}{\textbf{Macro F1 (\%)}} \\
\cmidrule(lr){3-5} \cmidrule(lr){6-8}
& & Overall & Inner & Outer & Overall & Inner & Outer \\
\midrule
\multicolumn{8}{l}{\textit{Digits Corruption}} \\
\multirow{5}{*}{Early} & random & 77.98{\scriptsize$\pm$0.04} & 21.23{\scriptsize$\pm$2.16} & 82.74{\scriptsize$\pm$0.01} & 78.15{\scriptsize$\pm$0.01} & 17.41{\scriptsize$\pm$1.86} & 82.53{\scriptsize$\pm$0.05} \\
& start & 77.74{\scriptsize$\pm$0.10} & 23.32{\scriptsize$\pm$0.45} & 82.68{\scriptsize$\pm$0.25} & 77.76{\scriptsize$\pm$0.11} & 19.38{\scriptsize$\pm$0.67} & 82.33{\scriptsize$\pm$0.29} \\
& end & 77.85{\scriptsize$\pm$0.21} & 23.96{\scriptsize$\pm$0.73} & 82.61{\scriptsize$\pm$0.30} & 77.95{\scriptsize$\pm$0.24} & 20.35{\scriptsize$\pm$0.82} & 82.33{\scriptsize$\pm$0.22} \\
& middle & 78.12{\scriptsize$\pm$0.16} & 22.07{\scriptsize$\pm$1.77} & 82.45{\scriptsize$\pm$0.12} & 78.29{\scriptsize$\pm$0.12} & 18.45{\scriptsize$\pm$1.23} & 82.24{\scriptsize$\pm$0.07} \\
& syntax & 77.92{\scriptsize$\pm$0.05} & 22.63{\scriptsize$\pm$0.84} & 82.76{\scriptsize$\pm$0.04} & 78.01{\scriptsize$\pm$0.06} & 18.95{\scriptsize$\pm$0.42} & 82.52{\scriptsize$\pm$0.10} \\
\midrule
\multirow{5}{*}{Late} & random & 76.72 & 17.43 & 81.60 & 77.04 & 14.75 & 81.50 \\
& start & 75.69 & 18.69 & 81.85 & 75.80 & 15.12 & 81.52 \\
& end & 75.51 & 13.28 & 82.31 & 75.65 & 11.59 & 81.98 \\
& middle & 77.30 & 17.87 & 82.73 & 77.41 & 15.64 & 82.62 \\
& syntax & 76.31 & 23.14 & 82.64 & 76.35 & 18.85 & 82.37 \\
\midrule
\multicolumn{8}{l}{\textit{Letters Corruption}} \\
\multirow{5}{*}{Early} & random & 78.08{\scriptsize$\pm$0.09} & 23.80{\scriptsize$\pm$2.34} & 82.10{\scriptsize$\pm$0.15} & 78.19{\scriptsize$\pm$0.06} & 19.77{\scriptsize$\pm$2.03} & 81.83{\scriptsize$\pm$0.12} \\
& start & 78.11{\scriptsize$\pm$0.12} & 22.51{\scriptsize$\pm$1.44} & 82.15{\scriptsize$\pm$0.25} & 78.24{\scriptsize$\pm$0.16} & 19.03{\scriptsize$\pm$1.13} & 81.88{\scriptsize$\pm$0.27} \\
& end & 77.81{\scriptsize$\pm$0.15} & 25.92{\scriptsize$\pm$0.24} & 82.39{\scriptsize$\pm$0.21} & 77.92{\scriptsize$\pm$0.11} & 21.54{\scriptsize$\pm$0.43} & 82.19{\scriptsize$\pm$0.06} \\
& middle & 77.46{\scriptsize$\pm$0.23} & 22.82{\scriptsize$\pm$0.83} & 82.42{\scriptsize$\pm$0.24} & 77.57{\scriptsize$\pm$0.27} & 19.57{\scriptsize$\pm$0.87} & 82.33{\scriptsize$\pm$0.20} \\
& syntax & 77.57{\scriptsize$\pm$0.15} & 23.19{\scriptsize$\pm$1.20} & 82.55{\scriptsize$\pm$0.28} & 77.67{\scriptsize$\pm$0.20} & 19.70{\scriptsize$\pm$0.85} & 82.28{\scriptsize$\pm$0.27} \\
\midrule
\multirow{5}{*}{Late} & random & 77.39 & 17.82 & 82.86 & 77.68 & 14.82 & 82.85 \\
& start & 77.78 & 23.30 & 83.01 & 77.89 & 19.65 & 82.95 \\
& end & 76.29 & 14.61 & 83.09 & 76.61 & 12.33 & 82.92 \\
& middle & 77.30 & 16.93 & 82.53 & 77.36 & 13.28 & 82.29 \\
& syntax & 77.29 & 23.84 & 82.40 & 77.31 & 20.79 & 82.11 \\
\midrule
\multicolumn{8}{l}{\textit{Diglets Corruption (mixed digits+letters)}} \\
\multirow{2}{*}{Early} & start & 78.08 & 22.43 & 82.27 & 78.16 & 18.86 & 82.03 \\
& end & 77.55 & 22.37 & 82.83 & 77.60 & 18.50 & 82.62 \\
\midrule
\multirow{2}{*}{Late} & start & 77.04 & 20.20 & 81.74 & 77.23 & 16.34 & 81.44 \\
& end & 75.64 & 13.85 & 82.13 & 76.10 & 11.48 & 82.05 \\
\midrule
\multicolumn{8}{l}{\textit{Semicolon Corruption}} \\
\multirow{2}{*}{Early} & start & 74.59 & 19.16 & 77.13 & 75.11 & 16.36 & 77.45 \\
& end & 74.37 & 18.34 & 74.65 & 74.90 & 16.66 & 75.05 \\
\midrule
\multirow{2}{*}{Late} & start & 76.24 & 3.44 & 83.67 & 76.21 & 2.77 & 83.10 \\
& end & 76.58 & 6.08 & 83.43 & 76.62 & 5.66 & 83.00 \\
\midrule
\multicolumn{8}{l}{\textit{Comma Corruption}} \\
\multirow{2}{*}{Early} & start & 74.23 & 19.46 & 76.38 & 74.84 & 17.88 & 76.91 \\
& end & 73.93 & 18.86 & 73.00 & 74.38 & 17.02 & 73.53 \\
\midrule
\multirow{2}{*}{Late} & start & 76.79 & 5.05 & 83.33 & 76.73 & 3.72 & 82.88 \\
& end & 76.30 & 5.22 & 83.56 & 76.43 & 4.57 & 83.30 \\
\bottomrule
\end{tabular}
\caption{Full entity corruption results on NEREL test set. ``Early'' (early damage) refers to training on corrupted data and predicting on clean data; ``Late'' (late damage) refers to training on clean data and predicting on corrupted data. Positions: random, start, end, middle, syntax (syntactic root).}
\label{tab:full-corruption-results}
\end{table*}

\end{document}